\documentclass[10pt,twocolumn,letterpaper]{article}

\usepackage{cvpr}
\usepackage{times}
\usepackage{epsfig}
\usepackage{graphicx}
\usepackage{amsmath}
\usepackage{amssymb}

\usepackage{tabularx} 
\usepackage{multirow}
\usepackage{slashbox}
\usepackage{booktabs} 
\usepackage{algorithm}
\usepackage{algorithmic}
\usepackage{bm}
\usepackage{float}

\usepackage[breaklinks=true,bookmarks=false]{hyperref}

\cvprfinalcopy 

\def\cvprPaperID2043 
\def\httilde{\mbox{\tt\raisebox{-.5ex}{\symbol{126}}}}

\begin{document}

\title{Learning with Batch-wise Optimal Transport Loss for 3D Shape Recognition}

\author{First Author\\
Institution1\\
Institution1 address\\
{\tt\small firstauthor@i1.org}
\and
Second Author\\
Institution2\\
First line of institution2 address\\
{\tt\small secondauthor@i2.org}
}

\author{
	Lin Xu$^{1,2}$\footnote{Contact Author} \and
	Han Sun$^{1,2}$\and
	Yuai Liu$^{1,2}$\and
	$^1$Institute of Advanced Artificial Intelligence in Nanjing,   $^2$Horizon Robotics\\
{\tt\small \{lin01.xu, han.sun, yuai.liu\}@horizon.ai}
}

\maketitle

\begin{abstract}
	Deep metric learning is essential for visual recognition. The widely used pair-wise (or triplet) based loss objectives cannot make full use of semantical information in training samples or give enough attention to those {hard samples} during optimization. Thus, they often suffer from a slow convergence rate and inferior performance. 
	In this paper, we show how to learn an importance-driven distance metric via optimal transport programming from batches of samples. It can automatically emphasize hard examples and lead to significant improvements in convergence. We propose a new batch-wise optimal transport loss and combine it in an end-to-end deep metric learning manner. We use it to learn the distance metric and deep feature representation jointly for recognition. Empirical results on visual retrieval and classification tasks with six  benchmark datasets, i.e., MNIST,  CIFAR10, SHREC13, SHREC14,  ModelNet10, and ModelNet40, demonstrate the superiority of the proposed method.  It can accelerate the convergence rate significantly while achieving a state-of-the-art recognition performance. For example,   in 3D shape recognition experiments, we show that our method can achieve better recognition performance  within only $5$ epochs than what can be obtained by mainstream 3D shape recognition approaches after $200$ epochs. 
	
\end{abstract}

\begin{figure}[htb]
	\begin{center}
		\includegraphics[height=9cm,width=7cm]{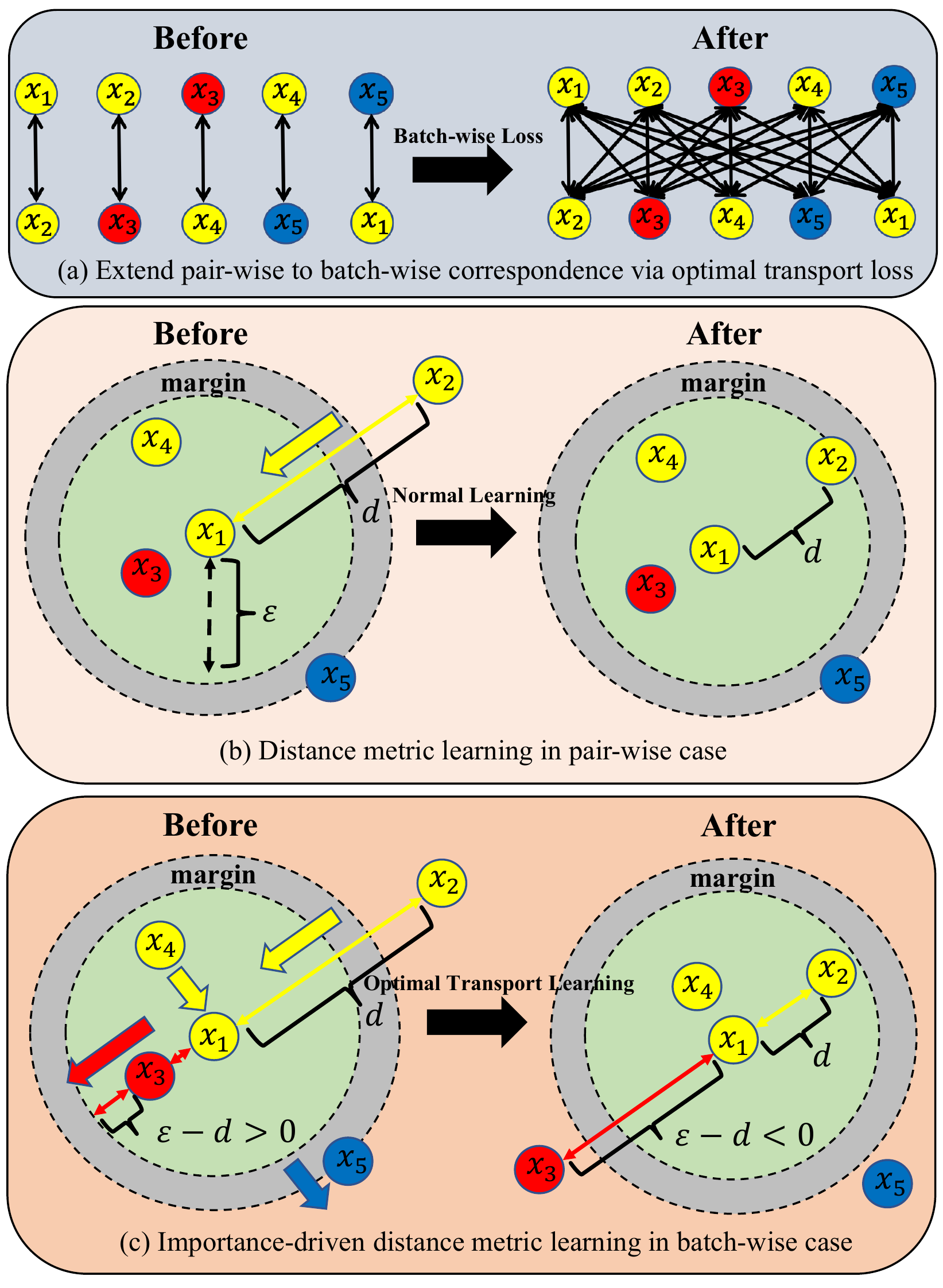}
	\end{center}
	\caption{Schematic illustration of learning with the proposed batch-wise loss objective as compared to pair-wise loss objective. The colors of circles represent semantical (or category) information.
		\textbf{(a):} The relationships among batches of samples of these two loss objectives. \textbf{(b):} Only the semantical information of a pair of examples is considered at each update. \textbf{(c):} The importance-driven distance metric is optimized using all available information within training batches so that similar positive examples with large ground distances and dissimilar negative examples with small ground distances are emphasized automatically. Arrows indicate the weights (or \emph{importance}) on distances arising from the proposed batch-wise optimal transport loss objective.}
	\label{fig:motivation}
	\vspace{-0.3cm} 
\end{figure}

\section{Introduction}
Learning a semantical embedding metric to make  similar positive samples cluster together, while  dissimilar negative ones widen apart is an essential part for modern recognition tasks \cite{hadsell2006dimensionality,chopra2005learning}. 
With the flourish of deep learning technologies \cite{krizhevsky2012imagenet,simonyan2014very,szegedy2015going}, deep metric learning has gained more attention in recent years \cite{hoffer2015deep,bell2015learning,schroff2015facenet,cui2016fine,sohn2016improved}. 
By training deep neural networks discriminatively end-to-end, a more complex highly-nonlinear deep feature representation (from the input space to a lower dimensional semantical embedding metric space) can be learned.
The jointly learned deep feature representation and embedding metric yield significant improvement for recognition applications, such as 2D image retrieval \cite{wang2014learning, bell2015learning, oh2016deep} or classification \cite{weinberger2009distance, qian2015fine}, signature verification \cite{bromley1994signature},  face recognition \cite{chopra2005learning, weinberger2009distance, schroff2015facenet},  and sketch-based 3D shape cross-modality retrieval \cite{li2013shrec, wang2015sketch, xie2017learning}.

Despite the progress made,  most of the pre-existing loss objectives  \cite{bromley1994signature,chopra2005learning,hoffer2015deep,schroff2015facenet,bell2015learning} do have some limitations for metric learning. 
Commonly used contrastive loss \cite{hadsell2006dimensionality,chopra2005learning} or triplet loss \cite{weinberger2009distance,chechik2010large}  only considers the semantical information within individual pairs or triplets of examples at each update,  while the interactions with the rest ones are ignored. It would bias the learned embedding metric and feature representation.
Moreover,  they do not give enough attention to hard positive or negative examples, by cause of the fact that these samples are often sparsely distributed and expensive to seek out. These \emph{hard samples} can strongly influence parameters during the network is learned to correct them. 
As a consequence,  methods which neglect them often suffer from slow convergence rate and suboptimal performance. 
Occasionally, such methods require expensive sampling techniques to accelerate the training process and boost the learning performance \cite{chechik2010large,schroff2015facenet,norouzi2012hamming,cui2016fine}.

In this paper,  we propose a novel batch-wise optimal transport loss objective for deep metric learning. It can learn an importance-driven distance metric via optimal transport programming from batches of samples simultaneously.  
As we know,  the fundamental idea behind metric learning is minimizing the intra-category variations (or distances) while maximizing the inter-category variations (or distances).  Thus, those semantically similar positive samples with large ground distances and dissimilar negative examples with small ground distances should be regarded as \emph{hard samples}.  Such samples should be emphasized correctly  to accelerate the metric learning process. 
Figure \ref{fig:motivation} illustrates our main idea of proposing  the new batch-wise optimal transport loss objective. As illustrated,  
learning with the proposed loss can utilize all available semantical information of training batches simultaneously.
The introduced importance-driven distance metric is partly obtained as a solution to the optimal transport program \cite{vallender1974calculation,cuturi2013sinkhorn}. It can  mine and emphasize those \emph{hard samples} automatically.
Thus,  the convergence rate  of  distance metric learning process can be significantly improved.
We further develop the new loss objective in a deep metric learning manner. 
The whole network can be trained discriminatively in an end-to-end fashion. The jointly learned semantical embedding metric and deep feature representation would be more robust to intra-class and inter-class variations. 
We finally verify the performance of our proposed method applying to various visual recognition tasks, including 2D image recognition, sketch-based 3D shape cross-modality retrieval, and 3D shape recognition.  
Experiment results on six widely used benchmark datasets, 
i.e.,  \emph{MNIST},  \emph{CIFAR10},  \emph{SHREC13}, \emph{SHREC14},   \emph{ModelNet10} and  \emph{ModelNet40},  demonstrate the superiority of the proposed method. 
Our method can achieve a state-of-the-art recognition performance with a notably fast convergence rate.


%
%
%


In a nutshell, our main contributions in the present work can be  summarized as follows:

(1) We propose a novel batch-wise optimal transport loss objective for learning an importance-driven distance metric to improve the existing pair-wise based loss objectives. 

(2) We develop a deep metric learning method based on the proposed loss objective, which learns the importance-driven metric and deep feature representation jointly.


(3) We verify the superiority of our proposed method on visual recognition tasks, including 2D image recognition,  sketch-based 3D shape retrieval, and 3D shape recognition.

%
%


\begin{figure*}[htb]
	\centering 
	\includegraphics[height=8.3cm,width=17.5cm]{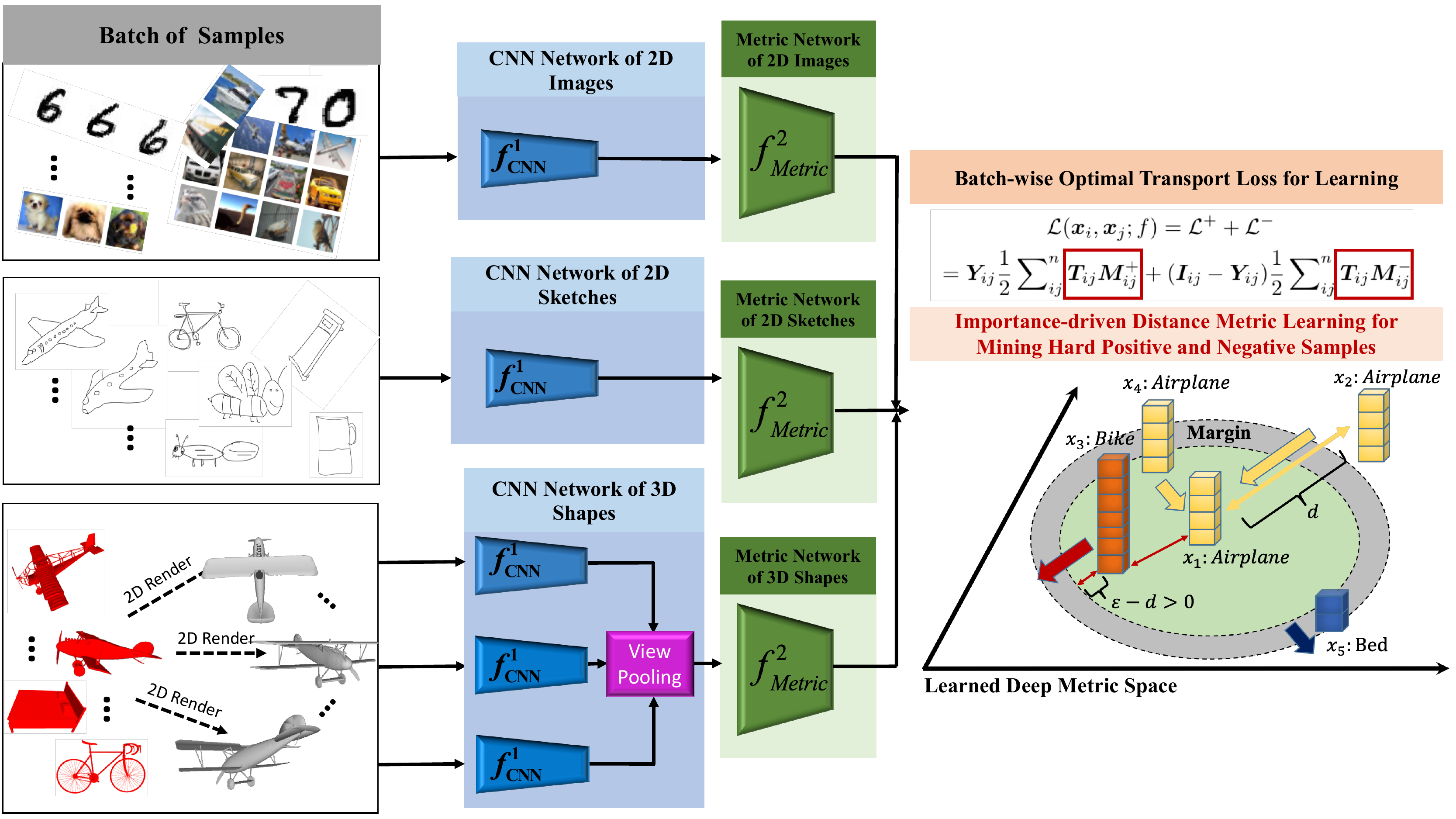}
	\caption{
		We formulate the proposed  loss into a deep metric learning framework. 
		Given batches of each modality samples, we use \emph{LeNet-5} \cite{lecun1998gradient}, \emph{ResNet-50} \cite{he2016deep}, and  \emph{MVCNN} \cite{su2015multi} as $\bm {{f^1_{\text{CNN}}}}$ to extract deep CNN features for 2D images, 2D sketches, and 3D shapes, respectively. The metric network $\bm {{f^2_{\text{Metric}}}}$ consisting of four fully connected (FC) layers, i.e., 4096-2048-512-128 (two  FC layers 512-256 for \emph{LeNet-5}) is used to perform dimensionality reduction of the CNN features.  We add three sigmoid functions as activation among these FC layers to generate normalized and dense feature vectors.  The whole framework can be  end-to-end trained discriminatively with the new batch-wise optimal transport loss.  The highlighted  
		importance-driven distance metrics
		${\bm T_{ij}}\bm M_{ij}^{+}$ and ${\bm T_{ij}}\bm M_{ij}^{-}$ are used for emphasizing {hard positive} and {negative samples}. It jointly learns the semantic embedding metric and deep feature representation for retrieval and classification.}
	\label{figFramework}
\end{figure*}

\section{Related Work}\label{sec2}
Recognition of 3D shapes is becoming prevalent with the advancement of modeling, digitizing, and visualizing techniques for 3D objects. 
The increasing availability of 3D CAD models,  both on the Internet, e.g.,  \emph{Google 3D Warehouse} \cite{GoogleWarehouse} and \emph{Turbosquid} \cite{Turbosquid},  and in the domain-specific field, e.g., \emph{ModelNet} \cite{ModelNet} and \emph{SHREC} \cite{li2015comparison},  
has led to the development of several scalable and efficient methods to study and analyze them, as well as to facilitate practical applications.
For 3D shape recognition, one fundamental issue is how to construct a determinative yet robust 3D shape descriptor and feature representation.  
Compared to 2D images, 3D shapes have more complex geometric structures.  Their appearance can be affected significantly by innumerable variations such as  viewpoint, scale,  and deformation. These have brought great challenges into the recognition task.
A natural method is to construct a shape descriptor based on the native 3D structures, e.g., point clouds, polygon meshes, and volumetric grid.  Then, shapes can be represented with distances, angles, triangle areas, tetrahedra volumes, local shape diameters \cite{osada2002shape,chaudhuri2010data},  heat kernel signatures \cite{bronstein2011shape,kokkinos2012intrinsic},  
extensions of  handcrafted SIFT, SURF \cite{knopp2010hough},  and learned 3D CNNs \cite{wu20153d,maturana2015voxnet} on 3D volumetric grids.
An alternative way is  describing a 3D shape by a collection of 2D view-based projections. 
It can make use of  CNN models, which have been pre-trained on  large 2D image datasets such as ImageNet \cite{krizhevsky2012imagenet} and gained a decent ability of generalization \cite{donahue2014decaf,simonyan2014very,girshick2014rich}. 
In this context, DeepPano \cite{shi2015deeppano} and PANORAMA-NN \cite{sfikas2017exploiting} are developed to convert
3D shapes into panoramic views, e.g., a cylinder projection around its principal axis. 
Multi-view CNN (\emph{MVCNN}) \cite{su2015multi} groups multiple CNNs with a view-pooling structure to process and learn from all available 2D projections of a 3D shape jointly.



\section{Background}\label{sec3}
\noindent \textbf{Loss objective for metric learning:}
Metric learning aims to learn a semantical metric  from input samples. 
Let $\bm x \in \bm X$ be an input sample. 
The kernel function $f(\cdot;\bm \theta):\bm X \to \mathbb{R}^d$ takes input $\bm{x}$ and generates an feature representation or embedding $f(\bm x)$. In deep metric learning \cite{hadsell2006dimensionality,chopra2005learning,sohn2016improved},  kernel $f(\cdot;\bm \theta)$  is usually defined by a deep neural network, parameterized by a series of weights and bias $\bm \theta$.  
Metric learning optimizes a discriminative loss objective to minimize  intra-class distances while maximizing  inter-class distances. 
For example,  the contrastive loss in seminal Siamese network \cite{hadsell2006dimensionality,chopra2005learning} takes pairs of samples as input and trains two identical networks to learn a deep metric $\bm M_{ij}$ as
\begin{equation}
\mathcal{L}({\bm x_i},{\bm x_j};f) = {\bm y_{ij}}\bm M_{ij} + (\bm 1 - {\bm y_{ij}})\max \{ 0,\varepsilon - \bm M_{ij} \}, 
\end{equation}
where the label $\bm y_{ij} \in \{0,1\}$ indicates whether a pair of $(\bm x_i,\bm x_j)$ 
is from the same class or not. The margin parameter $\varepsilon$ imposes a threshold of the distances among dissimilar samples.
Conventionally,  the Euclidian metric $\bm M_{ij}=||(f({\bm x_i}) - f({\bm x_j})||_2^2$ in the feature embedding space  is used to denote the  distance of a pair of samples.
Triplet loss \cite{weinberger2009distance,chechik2010large} shares a similar idea with contrastive loss, but extends a pair of samples to a triplet. For a given query $\bm x_i$, a similar sample $\bm x_j$ to the query $\bm x_i$, and a dissimilar one $\bm x_k$, the triplet loss can be formulated as 
\begin{equation}
\mathcal{L}(\bm x_i,\bm x_j,{\bm x_k};f) = \max \{ 0,{\bm M_{ij}} - {\bm M_{ik}} +  \varepsilon\}.
\end{equation}  
Intuitively, it encourages the distance between the dissimilar pair $\bm M_{ik}=||(f({\bm x_i}) - f({\bm x_k})||_2^2$ to be larger than the distance between the similar pair $\bm M_{ij}=||(f({\bm x_i}) - f({\bm x_j})||_2^2$ by at least a margin $\varepsilon$.






\noindent \textbf{Optimal transport distances:}
Optimal transport distances \cite{cuturi2013sinkhorn}, also known as Wasserstein distances \cite{vallender1974calculation} or Earth Mover's distances \cite{rubner2000earth}, define a distance between two probability distributions according to principles
from optimal transport theory \cite{villani2008optimal, wu2011witsenhausen}. 
Formally, let $\bm r$ and  $\bm c$ be  $n$-dimensional probability measures. The set of transportation plans between probability distributions $\bm r$ and $\bm c$
is defined as 
$
U(\bm r,\bm c): = \{ \bm T \in {\mathbb R_{+}^{n \times n}} | \bm{T1} = \bm r,\bm{T^T1} = \bm c \},
$
where $\bm 1$ is an all-ones vector.  
The set of transportation plans  $U(\bm r,\bm c)$ contains all nonnegative $n \times n$ elements with row and column sums $\bm r$ and $\bm c$,  respectively. 

Give an  $n \times n$ ground distance matrix $\bm M$, the cost of mapping $\bm r$ to $\bm c$ using a transport matrix $\bm T$ can be quantified as $\langle \bm{T,M}\rangle$, where $\langle.,.\rangle$ stands for the Frobenius dot-product. Then the problem defined in Equation (\ref{OT_problem})  
\begin{equation}\label{OT_problem} 
{D_{\bm M}}(\bm r, \bm c): = \mathop {\min }\limits_{\bm T \in U(\bm r, \bm c)} \langle \bm T, \bm M\rangle ,  
\end{equation} 
is called an optimal transport problem between $\bm r$ and $\bm c$ given ground cost $\bm M$. The optimal transport distance ${D_M}(\bm r, \bm c)$ measures the cheapest way to transport the mass in probability measure $\bm r$ to match that in $\bm c$.

Optimal transport distances define a more powerful cross-bin metric to measure probabilities compared with some commonly used bin-to-bin metrics, e.g., Euclidean, Hellinger, and Kullback-Leibler divergences. However, the cost of computing ${D_M}$ is at least $\mathcal{O}(n^3log(n))$ when comparing two $n$-dimensional probability distributions in a general metric space \cite{pele2009fast}. To alleviate it, Cuturi \cite{cuturi2013sinkhorn} formulated a regularized transport problem by adding an entropy regularizer to  Equation (\ref{OT_problem}).
This makes the objective function strictly convex and allows it to be solved efficiently.
Particularly, given a transport matrix $\bm T$, let $h(\bm T)=-\sum\nolimits_{ij}   {{\bm T_{ij}}\log {\bm T_{ij}}} $ be the entropy of $\bm T$. For any $\lambda >0$, the regularized transport problem can be defined as 
\begin{equation}\label{OT_problem_regularized} 
{D^{\lambda}_{\bm M}}(\bm r, \bm c): = \mathop {\min }\limits_{\bm T \in U(\bm r, \bm c)} \langle \bm T, \bm M\rangle- \frac{1}{\lambda }h(\bm T),  
\end{equation}        
where the lager $\lambda$ is, the closer this relaxation ${D^{\lambda}_M}(\bm r, \bm c)$ is to original ${D_{\bm M}}(\bm r, \bm c)$. Cuturi \cite{cuturi2013sinkhorn} also proposed the Sinkhorn's algorithm \cite{sinkhorn1967diagonal} to solve Equation (\ref{OT_problem_regularized}) for the optimal transport $\bm T^*$. Specifically, let the matrix $\bm K= exp(-\lambda \bm M)$ and solve it for the scaling vectors $\bm u$ and $\bm v$ to a fixed-point by computing $\bm u=\bm r./\bm{Kv}$, $\bm v= \bm c ./ \bm {K^Tu} $ in an alternating way. These yield the optimal transportation plan $\bm T^*=diag(\bm u) \bm K diag(\bm v)$. This algorithm can be solved with complexity $\mathcal{O}(n^2)$ \cite{cuturi2013sinkhorn}, which is significantly faster than exactly solving the original optimal transport problem.     

\section{Our Method}
In this section, we propose a deep metric learning scheme by using  principles of the optimal transport theory \cite{villani2008optimal}. 
Currently, research works with optimal transport distance \cite{cuturi2013sinkhorn,cuturi2014fast,cuturi2014ground}  mainly focus  on theoretical analysis and simulation verification.
Thus, it is hard to apply them into a large-scale 3D shape recognition contest directly.
To this end, we  have done the following three works to construct a trainable 
batch-wise optimal transport loss objective.   


%

\subsection{Importance-driven Distance Metric Learning}\label{Re-defined the ground distances}

Assuming  we are given two batches of samples, each batch has $n$ examples $\bm{X} \in \mathbb R^{d \times n}$. Let $\bm x_i \in \mathbb{R}^d$ be the representation of the $i^{th}$ shape.
Additionally, let $\bm r$ and $\bm c$ be the $n$-dimensional probability vectors for  two batches, where $r_i$ and $c_i$ denote the number of times shape $i$ occurs in $\bm r$ and $\bm c$ (normalized overall  samples in $\bm r$ and $\bm c$). The optimal transport   introduces a transportation plan $\bm T \in \mathbb{R}^{n \times n}$ such that $\bm T_{ij}$ describes how much of $r_i$ should be transported to $c_j$. 
As described in Equation (\ref{OT_problem_regularized}),  the optimal transport distance between batches $\bm r$ and $\bm c$ can be re-formulated as  
\begin{equation}\label{our_OTD}
\begin{aligned}
& \quad  \bm D^{\lambda}_{\text{OT}}(\bm r,\bm c) =  \mathop {\min }\limits_{\bm T \ge 0} \sum\nolimits_{i.j = 1}^n {{\bm T_{ij}}} \bm M_{ij}-\frac{1}{\lambda }h(\bm T_{ij}) \\ 
& \text{s.t.} \quad   \sum\nolimits_{j = 1}^n {\bm T_{ij} }  = \bm r  
\quad \! \text{and} \! \quad  \sum\nolimits_{i = 1}^n {{\bm T_{ij}} = \bm c \quad  \forall i,j.} 
\end{aligned}
\end{equation}
The learned optimal transportation plan $\bm T^*$ is  a probability distribution \cite{cuturi2013sinkhorn}, which aims to find the least amount of cost needed to transport the mass from  batch  $\bm r$ to  batch $\bm c$. The unit of cost corresponds to transporting a sample by the unit of ground distance. 
Thus, $\bm T^*$ solved by Equation (\ref{our_OTD})  prefers to assign higher importance values to samples with small ground distances while leaving fewer for others.

Utilizing such property, we define the importance-driven distance metric via imposing semantical information of samples.  
Specifically, we first define the ground distances for a pair of similar positive samples
as 
\begin{equation}\label{eq1}
\bm M^+ (\bm x_i,\bm x_j;f)=e^{- \gamma ||f(\bm x_i)- f(\bm x_j)||_2^2 },
\end{equation}
where $\gamma$ is a hype-parameter controlling the extent of rescaling.
This re-scaling operator shrinks  large Euclidian distance between similar samples.
After re-scaling  $\bm M^+$,  the learned $\bm T^*$ tends to put higher importance values on  those similar samples which have far Euclidian distances among each other (a.k.a.,  {\emph{hard postive samples}}), while putting lower on the others accordingly. Thus, it would accelerate the process that similar samples are getting close to each other.
For dissimilar negative samples, we  define the ground distances correspondingly as
\begin{equation}\label{eq2}
\bm M^-(\bm x_i,\bm x_j;f)=e^{- \gamma \text{max} \{0,\varepsilon -{ ||f(\bm x_i)- f(\bm x_j)||_2^2 }\}}. 
\end{equation}
The hinge loss  $\text{max} \{0,\varepsilon -{ ||f(\bm x_i)- f(\bm x_j)||_2^2 }\}$ penalizes the dissimilar samples within the margin $\varepsilon$ and ignores the others. Thus, contrary to the above similar samples case,
here the learned $\bm T^*$  will pay higher importance values on those dissimilar samples with small Euclidian distances (a.k.a., {\emph{hard negative samples}}), while assigning fewer on the others. Thus, it could accelerate the process that dissimilar samples are getting apart to each other.




\subsection{Batch-wise Optimal Transport Loss Learning}

Based on the defined distances $\bm M^+, \bm M^-$,  and  optimal transportation plan $\bm T^*$, now we can formulate a batch-wise optimal transport loss  for metric learning. It can  be viewed as an $n$-pairs extension version of the contrastive loss or triplet loss. 
We define the loss objective as
\begin{equation}\label{Eq:optimal transport loss}
\begin{aligned}
&  \quad  \quad  \quad \qquad \mathcal{L}(\bm x_i,\bm x_j;f) \! =\! \mathcal{L}^{+}+ \mathcal{L}^{-} \\
&=  \bm Y_{ij}\frac{1}{2}\sum\nolimits_{ij}^n {\bm T_{ij}}\bm M_{ij}^{+}\! + \! (\bm I_{ij}-\bm Y_{ij})\frac{1}{2}\sum\nolimits_{ij}^n {\bm T_{ij}}\bm M_{ij}^{-}, 
\end{aligned}  
\end{equation}
where $\bm Y_{ij}$ is a binary label assigned to a pair of training batches. Let $\bm Y_{ij}=1$ if sample $\bm x_i$ and $\bm x_j$ are deemed similar, and $\bm Y_{ij}=0$ otherwise. An all-ones matrix is denoted as $\bm I\in \mathbb{R}^{n \times n}$  and $n$ is the size of each training batch. 
In practice, ${\bm T_{ij}}\bm M_{ij}^{+}$ and ${\bm T_{ij}}\bm M_{ij}^{-}$ can be regarded as the importance-driven distance metric for  positive and negative samples, respectively. The optimal transportation plan ${\bm T^*}$  obtained by solving Equation (\ref{our_OTD}) is  a probability distribution of weights for emphasizing hard positive and negative samples during the loss objective optimization.
We just write the loss objective regarding only one pair of batches here for simplicity.  The overall data loss objective based on all training batches can be easily derived as $\sum \mathcal{L} $. 

\subsection{Batch Gradient Descent  Optimization}
We further derive the back-propagation form of the batch-wise optimal transport loss objective.  The proposed  loss objective can be embedded into a deep metric learning framework,  so that the whole network can be trained discriminatively end-to-end via batch gradient descent.

Since the batch-wise optimal transport distance is a fully connected dense matrix of pairs-wise ground distance, its gradient can be  deduced  as network flow manner.  Specifically,  we compute the gradient of  corresponding  loss $\mathcal{L}(\bm x_i,\bm x_j;f)$ with respect to embedding representations $f(\bm x_i)$ and $f(\bm x_j)$ as follows,
\begin{small}
	\begin{equation}\label{gradient}
	\begin{aligned}
	& \frac{{\partial \mathcal L}}{{\partial {f(\bm x_i)}}} \!= \! \sum\limits_{j = 1}^n {{\bm T^*_{ij}}({f(\bm x_i)} - {f(\bm x_j)})({\bm Y_{ij}} - (\bm I_{ij} - {\bm Y_{ij}}){\bm \delta _{ij}})} \\
	& \frac{{\partial \mathcal L}}{{\partial {f(\bm x_j)}}}\! = \! - \!\sum\limits_{i = 1}^n {{\bm T^*_{ij}}({f(\bm x_i)}\! - \!{f(\bm x_j)})({\bm Y_{ij}}\! -\! (\bm I_{ij}\! - \!{\bm Y_{ij}}){\bm \delta _{ij}})},
	\end{aligned}
	\end{equation} 
\end{small}where $\bm T^*$ is the optimizer obtained from  Equation (\ref{OT_problem_regularized}). 
Motivated by the fast optimal distance computation \cite{cuturi2013sinkhorn,cuturi2014fast,frogner2015learning}, we relax the linear program in Equation (\ref{our_OTD}) using the regularized entropy as in Equation (\ref{OT_problem_regularized}).
It allows us to approximately solve  Equation (\ref{OT_problem_regularized}) in $\mathcal{O}(n^2)$ time via $\bm T^*={diag} (\bm u) \bm K diag (\bm v)$, where $n$ is the size of batch.

The $\bm \delta$ here is also a binary indicator assigned to the pairs.   Let $\bm \delta_{ij}=1$ when the Eucildian distance between  shape $\bm x_i$ and $\bm x_j$ is within the margin (i.e., $\varepsilon - ||f(\bm x_i) - f(\bm x_j)||_2^2 > 0$), and $\bm \delta_{ij}=0$ otherwise.          
The  $f(\bm x_i)$ and $f(\bm x_j)$ are feature representations obtained through  deep neural networks.  Therefore, the gradient with respect to the network
can be  computed easily with the chain-rule in a back-propagation fashion,  
as far as $\frac{{\partial \mathcal L}}{{\partial {f(\bm x_i)}}}$ and $\frac{{\partial \mathcal L}}{{\partial {f(\bm x_j)}}}$ are derived. 
We also note that the defined ground distance $\bm M^+$ and $\bm M^-$  are just used to determine the optimal transportation plan $\bm T^*$ for re-weighting the importance of similar positive and dissimilar negative samples. We do not consider them as  variables to compute gradient in Equation (\ref{gradient}) for gradient updating. 



\section{Experiments}\label{sec4}


In this section, we evaluated the performance of the proposed method with applications to 2D image recognition (i.e., retrieval and classification), sketch-based 3D shape retrieval, and 3D shape recognition tasks. Six widely used benchmark datasets were employed in our experiments, including \emph{MNIST} \cite{lecun1998gradient},  \emph{CIFAR10} \cite{krizhevsky2014cifar}, \emph{SHREC13} \cite{li2013shrec}, \emph{SHREC14} \cite{pickup2014shrec}, \emph{ModelNet10}, and \emph{ModelNet40} \cite{wu20153d}.


\begin{figure*}[htb]
	\centering 
	\includegraphics[width=\linewidth]{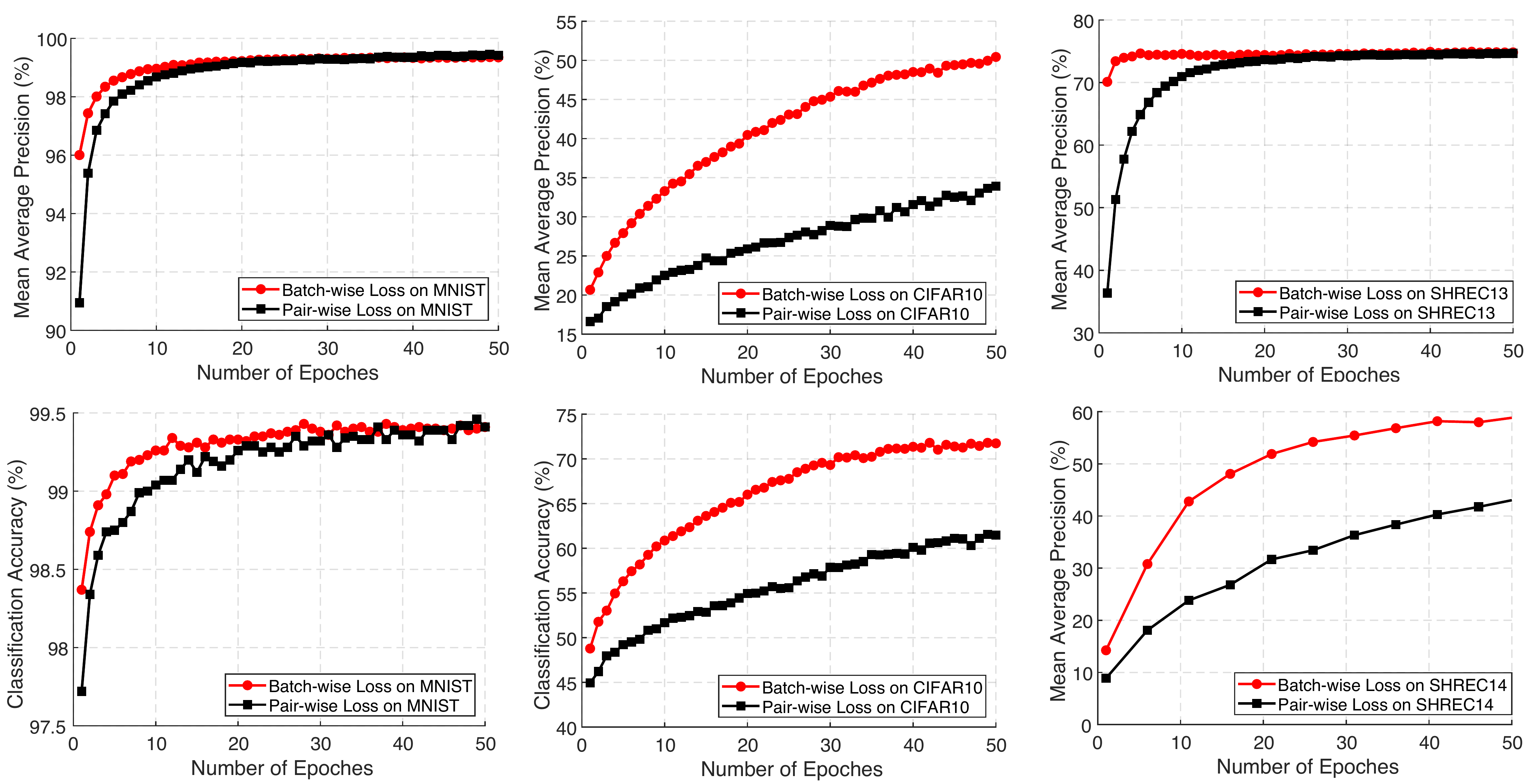}
	\caption{ \emph{Left:} Mean average precision (mAP) and classification accuracy curves of batch-wise optimal transport loss and pair-wise contrastive loss on 2D \emph{MNIST} dataset.  \emph{Middle:} Comparison of their mAP and accuracy curves on 2D \emph{CIFAR10} dataset.  \emph{Right:}  Comparison of their mAP curves on sketch-based 3D shapes \emph{SHREC13} and \emph{SHREC14} dataset.}
	\label{2dcurves}
\end{figure*} 
\begin{table*}
	\parbox{.5\linewidth}{
		\centering
		\caption{Retrieval results on the \emph{SHREC13} benchmark dataset}\label{T2d3d1} 
		\scalebox{.9}[1]{
			\begin{tabular}{c c c c c c c} 
				\hline
				\textbf{Method} & NN & FT & ST & E & DCG & mAP \\ [0.5ex] 
				\hline 
				CDMR & 0.279 & 0.203 & 0.296 & 0.166 & 0.458 & 0.250 \\ 
				\hline
				SBR-VC & 0.164 & 0.097 & 0.149 & 0.085 & 0.348 & 0.116 \\
				\hline
				SP & 0.017 & 0.016 & 0.031 & 0.018 & 0.240 & 0.026 \\
				\hline
				FDC & 0.110 & 0.069 & 0.107 & 0.061 & 0.307 & 0.086 \\
				\hline
				Siamese & 0.405 & 0.403 & 0.548 & 0.287 & 0.607 & 0.469\\ 
				\hline 
				LWBR & {0.712} &  {0.725} &  {0.725} &  \textbf{0.369} &  {0.814} & {0.752} \\
				\hline
				\textbf{Our Method} & \textbf{0.713} &  \textbf{0.728} &  \textbf{0.788} &  {0.366} &  \textbf{0.818} & \textbf{0.754} \\
				\hline
			\end{tabular}
		}
	}
	\hfill
	\parbox{.5\linewidth}{
		\centering
		\caption{Retrieval results on the \emph{SHREC14} benchmark dataset}\label{T2d3d2} 
		\scalebox{.9}[1]{
			\begin{tabular}{c c c c c c c} 
				\hline
				\textbf{Method} & NN & FT & ST & E & DCG & mAP \\ [0.5ex] 
				\hline 
				CDMR & 0.109 & 0.057 & 0.089 & 0.041 & 0.328 & 0.054 \\ 
				\hline
				SBR-VC & 0.095 & 0.050 & 0.081 & 0.037 & 0.319 & 0.050 \\
				\hline
				DB-VLAT & 0.160 & 0.115 & 0.170 & 0.079 & 0.376 & 0.131 \\
				\hline
				Siamese & 0.239 & 0.212 & 0.316 & 0.140 & 0.496 & 0.228\\ 
				\hline
				DCML & 0.272 & 0.275 & 0.345 & 0.171 & 0.498 & 0.286\\ 
				\hline 
				LWBR & {0.403} &  {0.378} &  {0.455} &  {0.236} &  {0.581} & {0.401} \\
				\hline
				\textbf{Our Method} & \textbf{0.536} &  \textbf{0.564} &  \textbf{0.629} &  \textbf{0.305} &  \textbf{0.712} & \textbf{0.591} \\
				\hline
			\end{tabular}
		}
	}
\end{table*}
\begin{figure*}[htb]
	\centering 
	\includegraphics[height=5.4cm,width=14cm]{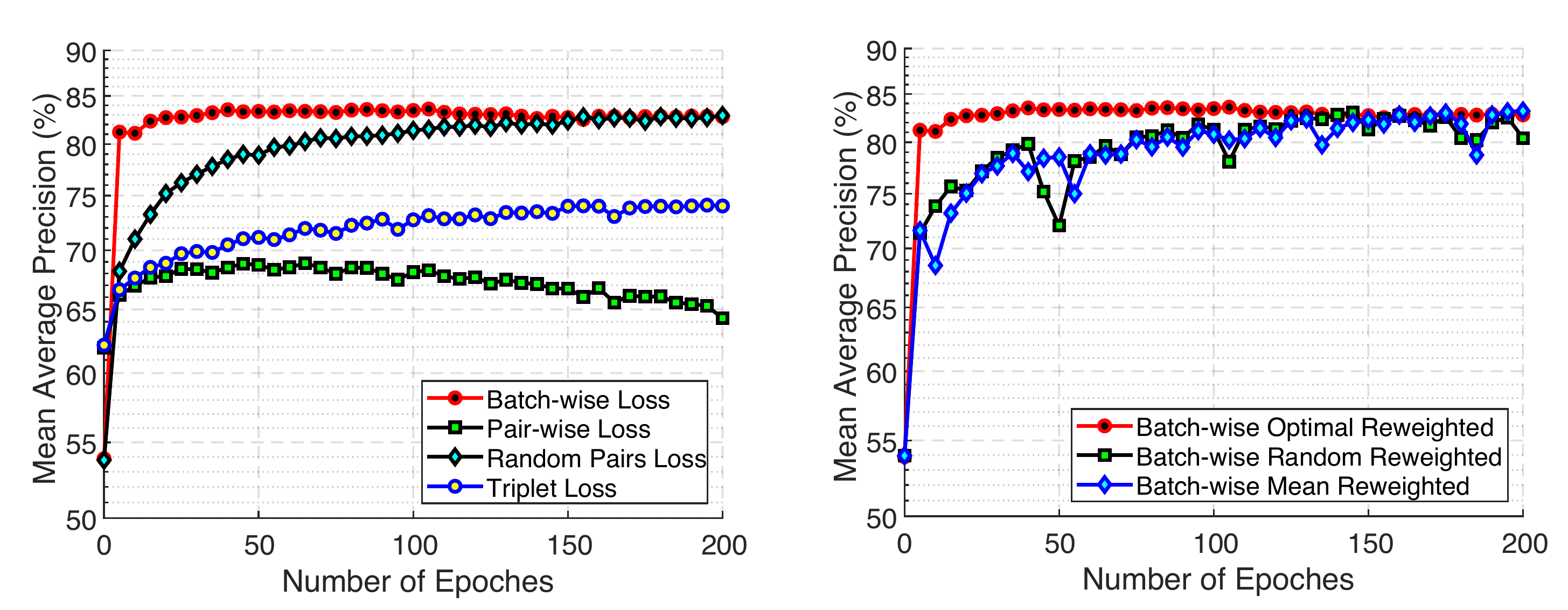}
	\caption{Mean average precision (mAP) curve with respect to the number of epochs evaluated of various methods on the \emph{ModelNet40} dataset. Left subfigure illustrates the mAP curves of four loss objectives for 3D shape retrieval, and right subfigure illustrates the mAP curves of three weighting modesc. The mAP have been observed every five epochs for 200 epochs in figures.}  
	\label{ow_curve}
\end{figure*}
\begin{table*}[htb]
	\renewcommand{\arraystretch}{1}
	\begin{center}
		\caption{Comparisons of batch-wise optimal transport loss with other benchmark methods on the \emph{ModelNet40} dataset.}\label{table1}
		\scalebox{.88}[.9]{
			\begin{tabular}{c c c c c c c c c} \toprule        
				~ & \multicolumn{8}{c}{ \textbf {Evaluation Criteria} } \\
				\cmidrule{3-9}\morecmidrules\cmidrule{3-9}
				&\textbf{ Methods}    & NN        & FT      & ST      & DCG    &   E   &  mAP (\%)  &  Accuracy (\%)  \\ \midrule
				\multirow{3}[0]*{\textbf{Pair-wise}}      
				& Individual      & 0.8287    & 0.6544    & 0.7891    & 0.8562        & 0.5668      & 69.3\%    &  88.6\% \\      
				& Triplets       & 0.8324    & 0.6968    & 0.8029    & 0.8629        & 0.5927      & 74.1\%    &  89.1\% \\            
				& Random     & 0.8688    & 0.7948     & 0.9048    & 0.9140         & 0.6601      & 83.1\%  &  89.5\%\\             
				\multirow{3}[0]*{\textbf{Batch-wise}}
				& Mean Weighted        & 0.8750    & 0.7986     & 0.9032    & 0.9158        & 0.6589         & 83.3\%         &  89.7\%     \\      
				& Random Reweighted      & 0.8688    & 0.7673     & 0.8846    & 0.9051         & 0.6445      & 83.1\%         &  89.0\%     \\ 
				& \textbf{Optimal Reweighted}     & \textbf{0.8762 }   &\textbf{0.8013}    & \textbf{0.8991}  & \textbf{0.9178 } &\textbf{0.6560}   & \textbf{83.8 \%}  &  \textbf{90.3 \% }  \\ 
				\bottomrule                      
		\end{tabular} }
	\end{center}
	\vspace{-0.5cm}
\end{table*}
\begin{table*}[!htb]
	\renewcommand{\arraystretch}{1}
	\begin{center}
		\caption{Retrieval and classification results on the \emph{ModelNet10} and \emph{ModelNet40} datasets.}\label{table2}
		\scalebox{.9}[.9]{
			\begin{tabular}{l l c c c c}\toprule
				\textbf{Methods} & \textbf{Shape Descriptor} & \multicolumn{2}{c}{\textbf{\emph{ModelNet10}}} & \multicolumn{2}{c}{\textbf{\emph{ModelNet40}}} \\
				\cmidrule{3-6}\morecmidrules\cmidrule{3-6}   
				&  &  mAP (\%)  &  Accuracy (\%) &  mAP (\%)  &  Accuracy (\%)  \\ \midrule
				(1) MVCNN   \cite{su2015multi}  & 2D View-based Descriptor (\#Views=12)   & N/A & N/A  & 80.2\%      & 89.5\%  \\
				& 2D View-based Descriptor (\#Views=80)  & N/A & N/A  & 79.5\%      & 90.1\%  \\ \midrule
				(2) GIFT \cite{bai2016gift}     & 2D View-based Descriptor (\#Views=64)   &\textbf{91.1 \%} & 92.3\%  & 81.9\%      & 83.1\%  \\  \midrule                       
				(3) 3DShapeNets \cite{wu20153d} & 3D Voxel Grid ($30\times30\times30$)    &68.3\%  & 83.5\%  & 49.2\%      & 77.0\%  \\\midrule     
				(4) Geometry Image \cite{sinha2016deep}      & 2D Geometry Image   & 74.9\% &  88.4\%   & 51.3\%      &83.9\% \\\midrule                                                       
				(5) PANORAMA-NN \cite{sfikas2017exploiting}  & 2D  Panoramic View  & 87.4\% &  91.1\%    & 83.5\%      &\textbf{90.7 \% } \\\midrule      
				(6) DeepPano \cite{shi2015deeppano}          & 2D  Panoramic View    & 84.1\% & 85.4\%  & 76.8\%      &77.6\%  \\\midrule     
				\textbf{Our Method}             & 2D View-based Descriptor (\#Views=12)  & 87.5\%    & \textbf{93.7 \%}  & \textbf{83.8 \%}      & 90.3 \%  \\        
				\bottomrule
		\end{tabular}}
	\end{center}
	\vspace{-0.4cm}
\end{table*}
\begin{figure}[htb]
	\centering
	\includegraphics[width=\linewidth]{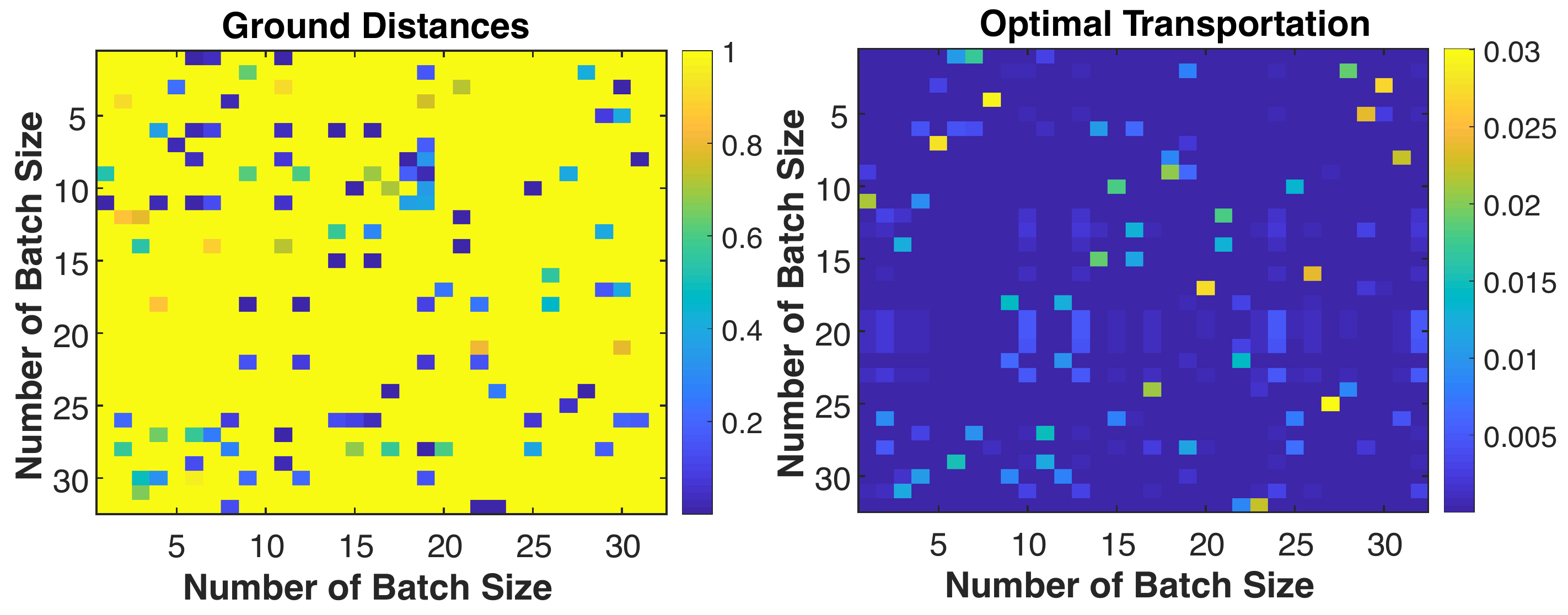}
	\caption{Illustration of  relationship between the ground distances $\bm M^*$ and the optimal transportation plan $\bm T^*$ on the \emph{ModelNet40} dataset. 
		For two batches (each with batch size $32$) of samples, we visualize the values of the batch-wise ground distances matrix (i.e., $32 \times 32$)  and the corresponding optimal transportation plan. }
	\label{fig:distance}
\end{figure}

\subsection{Experimental settings}

\noindent \textbf{Architecture:}  Figure \ref{figFramework} illustrates network architecture of deep metric learning with our batch-wise loss objective.

\noindent \textbf{Datasets:}
The \emph{MNIST} \cite{lecun1998gradient} is a large handwritten digits  dataset, which has $60,000$ $28 \times 28$ black-and-white training images and 10,000 testing images. 
The \emph{CIFAR10}  \cite{krizhevsky2014cifar} dataset  consists of  $60,000$ $32 \times 32$ RGB images in 10 different categories, with $6,000$ images per category. There are $50,000$ training images and $10,000$ test images. 
\emph{SHREC13} \cite{li2013shrec} and  \emph{SHREC14} \cite{pickup2014shrec} are two large-scale datasets for sketch-based 3D shape retrieval. 
\emph{SHREC13} contains  $7,200$ human-drawn sketches and $1,258$ 3D shapes from 90 different categories.  For each category,  $50$ sketches are used for training and remaining $30$ sketches are used for the test. There are $14$ 3D shapes per category generally. \emph{SHREC14}
is larger than \emph{SHREC13}, which has $13,680$ sketches and $8,987$ 3D shapes from $171$ categories.  Each of the categories has $53$ 3D shapes on average. There are $8,550$ sketches for training and $5,130$ for test.
\emph{ModelNet} \cite{ModelNet} is a large-scale 3D shape dataset, which contains $151,128$ 3D CAD models belonging to $660$ unique object categories \cite{wu20153d}. There are two subsets of \emph{ModelNet} can be used for evaluation. \emph{ModelNet10} contains $4,899$ 3D shapes from $10$ categories while \emph{ModelNet40} has $12,311$ shapes from 40 categories. In our experiments, we used
the same training and test splits as in \cite{wu20153d}. Specifically, we randomly selected $100$ unique shapes per category, where $80$ shapes were chosen for training and the remaining $20$ shapes for the test. 


\noindent \textbf{Evaluations:} 
For retrieval, we used Euclidian distance to measure the similarity of the shapes based on  their learned feature vectors output by the metric network as shown in Figure \ref{figFramework}.
Given a query from the test set, a ranked list of the remaining test samples was returned according to their distances to the query sample. 
We used the evaluation metrics for retrieval as in \cite{xie2017learning} when presenting our results. The metrics include nearest neighbor (NN) \cite{cover1967nearest}, first tier (FT) \cite{tangelder2004survey}, second tier (ST) \cite{cornea20053d}, E-measure (E) \cite{chen2004adaptive}, discounted cumulated gain (DCG) \cite{jarvelin2008discounted}, 
and mean average precision (mAP) \cite{philbin2007object}. 
For classification, we trained  one-vs-all linear SVMs \cite{chang2011libsvm} to classify 2D images and 3D shapes
using their features. The average category accuracy \cite{wu20153d} was used to evaluate the classification performance.


\noindent \textbf{Parameters settings:}
In our 2D image recognition, the learning rate and batch size  were $0.01$ and $64$ respectively. Our optimizer had a momentum of $0.9$ and $0$ weight decay rate. The regularized parameter $\lambda$ in Equation (\ref{our_OTD}) was set to be $0.01$ while 
the re-scaling parameter $\gamma$ in Equation (\ref{eq1}) being $10$. 
In the sketch-based 3D shape retrieval and 3D shape recognition experiments, the batch size was reduced to $32$. Meanwhile,  the learning rate, weight decay and momentum remained the same as what has been used in 2D experiments. We increased the regularized  parameter $\lambda$  to $10$,  which is the same as the re-scaling parameter $\gamma$.



\subsection{Evaluation of Proposed Method}




\subsubsection{2D Image Recognition}

Firstly, we empirically evaluated the effect of our proposed method on two broadly used 2D images benchmark datasets, i.e., \emph{MNIST} and  \emph{CIFAR10}.  
As illustrated in Figure \ref{figFramework}, we used a \emph{Siamese}-like symmetrical network structure, which employed \emph{Lenet-5} as its base CNN to obtain 
$256$-dimensional feature vectors
for the images in both datasets. Training images were randomly shuffled at the start of each epoch. In each training step, the optimal transportation $\bm T^*$ between two batches of image features was approximated by iterating the Sinkhorn's algorithm for 20 times. After each epoch, we computed all image features with the symmetrical network trained so far for classification and retrieval. The categorical accuracies provided by one-vs-rest linear SVMs and the retrieval mAPs given by the similarity measure based on the testing samples were recorded.



The left-hand and {middle} subfigures in Figure \ref{2dcurves} present accuracy and mAP curves of the batch-wise optimal transport loss learning concerning the number of epochs. These figures illustrate the relationship between the convergence rate and recognition performance.
Comparing with the pair-wise contrastive loss, our method posses a significantly faster convergence rate. On CIFAR10, it provides a retrieval mAP and a classification accuracy which are approximately $15 \%$ and $10 \%$ higher than the corresponding values achieved by the pair-wise loss at the end of $50$ epochs.
The empirical results indicate that the importance-driven distance metric learning can
effectively adjust the distribution of weights. It pays more attention to the hard positive and negative samples  during the training process.

%
%
\subsubsection{Sketch-based 3D Shape Retrieval}

We then evaluated our method for sketch-based 3D shape retrieval on two large-scale benchmark datasets, i.e.,  \emph{SHREC13} and \emph{SHREC14}. 
The right-hand two subfigures in Figure \ref{2dcurves}  demonstrate the mAP curves of our batch-wise optimal transport loss as compared to the pair-wise loss objective.  As illustrated, our method is about $5$ times and $3$ times faster than LBWR on \emph{SHREC13} and \emph{SHREC14} respectively. Meanwhile, the retrieval performance is remarkably higher than the compared LBWR.


We also compared our method with several mainstream approaches for 3D shape retrieval, including CDMR \cite{furuya2013ranking}, SBR-VC \cite{li2013shrec}, SP \cite{sousa2010sketch}, FDC \cite{sousa2010sketch}, Siamese network \cite{wang2015sketch}, DCML \cite{dai2017deep}, DB-VLAT \cite{tatsuma2012large},  and LWBR \cite{xie2017learning}. 
The evaluation criteria include  NN, FT, ST, E, DCG, and mAP.

As summarized in Table \ref{T2d3d1} and Table \ref{T2d3d2}, our batch-wise optimal transport loss  based method achieved the best retrieval performance with respect to all evaluation metrics on  \emph{SHREC13} and \emph{SHREC14}.
Among  compared methods, CDMR, DCML, Siamese network, and LWBR are all deep metric learning based approaches. They measured similarity based on pairs of samples and mapped data into an embedding metric space through different pooling schemes. In contrast,
our proposed batch-wise optimal transport loss objective can correctly re-weight the importance value of samples, mainly focus on those \emph{hard samples}.  Thus, our approach obtained better retrieval performance. Its mAP reaches 0.754, which is slightly better than LBWR and significantly better than other methods. Furthermore, the advantage of our approach is enlarged on \emph{SHREC14} because this dataset has more severe intra-class and cross-modality variations. As a consequence, the mAP of our proposed method is $0.591$, which is $0.190$, $0.305$,  and 0.363 higher than LBWR, DCML and Siamese network, respectively.

\subsubsection{3D Shapes Recognition}

We finally verified the proposed method for  3D shape recognition on two large-scale 3D shape datasets, i.e., \emph{ModelNet10} and \emph{ModelNet40}.  Pair-wise loss and triplet loss suffer from slow convergence rate because they are not capable of exploring all available semantical information within training batches simultaneously. To alleviate this problem, we used random sampling techniques (i.e., recurrently shuffle the training batches during each epoch) to loop over as many randomly sampled pairs as possible.
It is expected that the random pairs based loss objective could make full use of all information so that the finally learned semantic metric could be balanced correctly.  
The left-hand subfigure in Figure \ref{ow_curve} presents the mAP curves of batch-wise optimal transport loss and other compared loss objectives for 3D shape retrieval. Similarly,  
the batch-wise optimal transport loss objective still has significantly faster convergence rate and can achieve a decent retrieval performance within a small number of epochs (i.e., $5$ epochs).


We also examined two different probability distributions, i.e., uniformly distributed mean value ($\nu=\frac{1}{{{n^2}}}$) and random numbers in the interval $(0,1)$,   
as  alternatives of the optimal transportation plan. 
Uniformly distributed mean value weights in the batch-wise loss imply that samples  are equally important for later metric learning.  Uniformly distributed random weights randomly mark some samples as \emph{hard samples} within a pair of batches during the learning process.
The right-hand subfigure in  Figure \ref{ow_curve}  illustrates comparison results of the retrieval performance concerning the number of epochs for these three re-weighting strategies. It demonstrates that  the convergence rate of optimal re-weighted is much faster than the others. 


The detailed comparison results are summarized in Table \ref{table1}. We compared the batch-wise optimal transport loss with other designed benchmark methods using NN, FT, ST, E, DCG and mAP on the \emph{ModelNet40} dataset. As illustrated  in Figure \ref{ow_curve} and  Table \ref{table1}, learning with batch-wise optimal transport loss objective has considerably faster convergence rate than other benchmark methods. It  takes only a few epochs (i.e., $5$ epochs) to achieve mAP at  $83.8 \%$ and accuracy at $90.3 \%$, which are better than  others after $200$ epochs.
It demonstrates that the learned optimal transportation plan can correctly re-weight the training samples according to their importance values during the metric learning process. Moreover,  solving Equation (\ref{our_OTD}) to learn optimal transportation plan ${\bm T^*}$ is not computational expensive in practice. 
The average running time required by
one epoch of individual pair loss objective is $2.51$ seconds, and that of batch-wise optimal transport loss objective takes $9.02$ seconds.      

Here, we analyzed the role of learned importance-driven distance metric $\bm T^* \bm M^*$ in our work. It is composed of the optimal transportation plan $ \bm T^* \in \mathbb{R}^{n \times n}$ and the semantical information embedded ground distances $\bm M^* \in \mathbb{R}^{n \times n}$. The element $\bm M_{ij}^*$ is filled with the distances of batch-wise similar positive samples $\bm M_{ij}^+$ and the distances of batch-wise dissimilar negative samples $\bm M_{ij}^-$. 
The right-hand subfigure in Figure \ref{fig:distance} shows that  far similar positive and adjacent dissimilar negative samples (i.e., \emph{hard samples}) are sparsely distributed. 
The left-hand subfigure  is the optimal transportation plan which is actually a probability distribution  \cite{cuturi2013sinkhorn}. 
The color map reveals the learned metric ensures 
higher importance weights to those few samples with small ground distance while giving less on the remaining ones.



In the end, we compared our method with state-of-the-art approaches for shape retrieval and classification, including  MVCNN \cite{su2015multi}, GIFT \cite{bai2016gift}, DeepPano \cite{shi2015deeppano} and et al. 
The detailed comparison results are summarized in Table \ref{table2}.  Compared to these approaches,  our method based on batch-wise optimal transport loss learning can achieve the (almost) state-of-the-art performance on both tasks.   



\section{Conclusion}\label{sec5}


In this paper, we proposed a novel batch-wise optimal transport loss objective to learn an importance-driven distance metric.  The learned distance metric can effectively emphasize \emph{hard samples} according to their importance weights.
We then formulated the proposed loss objective into an end-to-end deep metric learning network for recognition. 
We evaluated the performance and versatility of our method with various visual recognition tasks, including 2D image recognition, 2D sketch-based 3D shape cross-modality retrieval, and multiview based 3D shape recognition.  
The empirical results verified the proposed method could generically accelerate the convergence rate while achieving excellent recognition performance. 
Our future work will involve facilitating such a trend and applying this importance-driven distance metric learning to more widespread applications.
For example, 3D point cloud classification,  segmentation,  3D scene reconstruction,   cross-modality correspondence among visual, audio,  and text.




%



{\small
	\bibliographystyle{ieee}
	\bibliography{egbib2}
}

\end{document}